# Rethinking Tokenization: Crafting Better Tokenizers for Large Language Models

Jinbiao Yang

Jinbiao.Yang@mpi.nl

Language and Computation in Neural Systems Group, Max Planck Institute for Psycholinguistics

**Abstract**

Tokenization significantly influences language models(LMs)' performance. This paper traces the evolution of tokenizers from word-level to subword-level, analyzing how they balance tokens and types to enhance model adaptability while controlling complexity. Despite subword tokenizers like Byte Pair Encoding (BPE) overcoming many word tokenizer limitations, they encounter difficulties in handling non-Latin languages and depend heavily on extensive training data and computational resources to grasp the nuances of multiword expressions (MWEs). This article argues that tokenizers, more than mere technical tools, should drawing inspiration from the cognitive science about human language processing. This study then introduces the "Principle of Least Effort" from cognitive science, that humans naturally seek to reduce cognitive effort, and discusses the benefits of this principle for tokenizer development. Based on this principle, the paper proposes that the Less-is-Better (LiB) model could be a new approach for LLM tokenizer. The LiB model can autonomously learn an integrated vocabulary consisting of subwords, words, and MWEs, which effectively reduces both the numbers of tokens and types. Comparative evaluations show that the LiB tokenizer outperforms existing word and BPE tokenizers, presenting an innovative method for tokenizer development, and hinting at the possibility of future cognitive science-based tokenizers being more efficient.

# Introduction

When confronted with vast or intricate information, our brains typically simplify it into smaller, more digestible segments, thereby helping us better understand and remember. Language, exemplifying such complexity, often requires segmenting itself into "chunks" (Isbilen & Christiansen, 2020). In the field of natural language processing (NLP), the chunks are often referred to as *tokens* through the process known as *tokenization*.

The choice of tokenizer has a crucial impact on the performance of language models. Especially in language models (LMs), how a tokenizer segments corpora determines the fundamental way the model processes language. This article investigates the roles of tokens (the actual number of lexical units in a corpus) and types (the number of different lexical units of vocabulary) in tokenizer design, and attempts to find an ideal solution that optimizes the number of tokens while controlling the number of types. In the following sections, the article will explore the advantages and limitations of subword tokenizers, analyze the treatment of Multiword Expressions (MWE) in current large language models. This article also argues that tokenizers, more than mere technical tools, should emulate and learn from human language processing methods, as tokenizers deal with content generated by humans rather than natural phenomena like sound and images, and calls for a general theory to guide the development of tokenizers. The article will discuss the "Principle of Least Effort" as such a general theory from cognitive science, and introduce a new type of tokenizer model - the *Less-is-Better* model, based on the Principle of Least Effort.

## From Word-level Tokenizers to Subword-level Tokenizers

NLP applications initially relied on word-level tokenizers, which divided text into words using spaces and punctuation. For example, the historical development of semantic representations started with the Bag-of-Words model, progressed to Word2Vec by (Mikolov et al., 2013), and to GloVe (Global Vectors for Word Representation) by (Pennington et al., 2014). They all aimed at training semantic representations at the level of words. Word-level tokenizers are relatively effective in processing European languages, where spaces provide clear word boundaries. However, this method is limited in languages like Chinese, which do not have clear word boundaries. Moreover, the flexible morphological inflections in language, the constant emergence of new words, and the prevalence of spelling errors in corpora make it difficult for word-level vocabularies to generalize in practical applications.

Subword technology can be traced back to the 1990s (Gage, 1994). Initially, these techniques were mainly used to compress data. With the emergence of large language models (LLMs), the demand for tokenizers increased. These complex models require an understanding and generation of extremely rich and diverse language content, and traditional word-level tokenizers struggle with complex vocabularies, morphological inflections, and the continuous influx of new vocabularies. At this point, subword-level tokenizers became the new mainstream due to their flexibility and generalization capabilities.

| Tokenizer | Representative Papers | Year | Used in Notable Models |
|---|---|---|---|
| BPE | "Neural Machine Translation of Rare Words with Subword Units" (Sennrich et al., 2016) | 2016 | GPT-2&3&4 |
| SentencePiece | "SentencePiece: A simple and language independent subword tokenizer and detokenizer for Neural Text Processing" (Kudo & Richardson, 2018) | 2018 | ALBERT (Lan et al., 2020), T5 (Raffel et al., 2019), XLNet (Z. Yang et al., 2020) |
| Unigram | "Subword Regularization: Improving Neural Network Translation Models with Multiple Subword Candidates" (Kudo, 2018) | 2018 | T5 (Raffel et al., 2019) |
| WordPiece | "Japanese and Korean Voice Search" (Schuster & Nakajima, 2012); "BERT: Pre-training of Deep Bidirectional Transformers for Language Understanding" (Devlin et al., 2019) | 2019 | BERT (Devlin et al., 2019), ERNIE (Sun et al., 2019) |

*Table 1: Popular tokenization methods that contributed to the evolution of language models in recent years.*

| Categories | Examples |
|---|---|
| English text | Generative Pre-trained Transformer 4 (GPT-4) is a multimodal large language model created by OpenAI, and the fourth in its series of GPT foundation models. |
| English subwords | Gener\|ative\| Pre\|-trained\| Transformer\| \|4\| (\|G\|PT\|-\|4\|)\| is\| a\| multim\|odal\| large\| language\| model\| created\| by\| Open\|AI\|,\| and\| the\| fourth\| in\| its\| series\| of\| G\|PT\| foundation\| models\|. |
| Chinese text | 您可以使用下面的工具了解语言模型如何对一段文本进行标记化，以及这段文本中的标记总数。 |
| Chinese subwords | 您\|可以\|使用\|下\|面\|的\|工\|具\|了\|解\|语\|言\|模\|型\|如\|何\|对\|一\|段\|文\|本\|进\|行\|标\|记\|化，\|以\|及\|这\|段\|文\|本\|中\|的\|标\|记\|总\|数\|。 |

*Table 2: Examples of text segmentation using BPE. The tokenizer used is provided officially by OpenAI for GPT-3.5 and GPT-4 [1].*

**Balancing Tokens and Types by Subwords**

In the transition from word-level to subword-level, a core consideration is how to balance the number of tokens and types. Word-level tokenizers, although producing fewer types, are unable to deal with Out-Of-Vocabulary (OOV) units. In contrast, subword-level tokenizers significantly reduce the occurrences of OOV, enhancing the model's adaptability to new vocabularies and complex language phenomena.

For example, BPE and WordPiece, in creating their vocabularies, effectively handle rare vocabularies by gradually merging frequently occurring character pairs or combinations, while keeping the number of tokens within a reasonable range. SentencePiece and Unigram models further improve adaptability to different languages, especially in languages that do not use spaces to separate words (like Chinese).

| Segmentation | #Tokens | #Types |
|---|---|---|
| Words | 100 million | 1,750,000 |
| BPE | 111 million | 82,000 |
| Characters | 550 million | 3,000 |

*Table 3: An example of the number of tokens/types with different tokenizers on a German corpus (See Table 1 in Sennrich et al., (2016)).*

As shown in the comparison in Table 3, the number of types in BPE is 4.7% of words, while the number of tokens is roughly equal (111%); the number of types in characters is 0.2% of words, but the number of tokens is 550%. This shows that the subword approach is high-yield (significantly reduced number of types) and low-cost (slightly higher number of tokens).

For languages with rich morphology (like the German corpus shown in Table 3), the significant reduction in the number of types with subword-level tokenization is mainly because it can break down words into frequent subunits, capturing morphological variations without needing separate entries for each word form. Although Chinese does not have a lot of morphological variations, the presence of numerous compound words (composed of two or more morphemes, like "关闭" [shut down], "直升机" [helicopter]) means that subword-level tokenization can also reduce the number of types.

The reduction in types lowers the computational complexity and memory requirements of LMs, having a direct positive impact on the models' performance. Moreover, subword-level tokenization has a stronger generalization capability for OOV contents or spelling errors in the corpus, as shorter tokens have a higher probability to cover more of the corpus. Thus, LLMs generally adopt the subword approachs for alphabetic languages (Table 1).

---

[1] https://platform.openai.com/tokenizer

This shift from word-level to subword-level not only marks the progress of tokenizer technology in the NLP field but also reflects a deeper understanding of language diversity and complexity. However, as seen in the examples in Table 2, unlike languages using the Latin alphabet, Chinese BPE subwords are mostly single characters. An analysis[2] shows that a sentence in Chinese may require 1.7x more tokens than a similar sentence in English, and Burmese or Amharic may require 10x more tokens. This indicates that for LLMs in various languages, especially non-Latin ones (like Chinese), BPE subwords still have shortcomings. Furthermore, although subword tokenizers have made significant progress in handling OOV issues, they still have limitations in capturing the nuanced semantics and idiom implications of language. This leads to the need for direct handling of multiword expressions as a complement and refinement of current tokenizer technology.

**Current Marginalization of Multiword Expressions (MWEs) in Language Models**

Multiword Expressions, despite playing a crucial role in everyday language, are often overlooked in the development of LMs. So far, only AI21 Studio's Jurassic-X models (Lieber et al., 2021) has introduced multi-word tokens, including expressions, phrases, and named entities, into their vocabularies. This marginalization may be primarily due to several reasons:

1. **Performance and complexity considerations**: Introducing MWEs as independent tokens will obviously increase the number of types. The vocabulary of the aforementioned Jurassic model includes about 250,000 types, much larger than most existing vocabularies (5 times or more)[3]. However, MWEs can be rare or highly specific to certain contexts or domains, so their introduction as independent tokens does not significantly reduce the total number of tokens. This somewhat contradicts the goal of efficient performance pursued by LMs. Moreover, low-frequency MWEs lead to insufficient representation in the training data, making it difficult for the models to learn and accurately predict their semantics.

2. **The alternative role of big data and computational power**: Current LLMs, like GPT-series and BERT, rely on massive training data and high computational power to learn the real usages of MWEs rather than their literal expressions, even though these expressions are not treated as whole units during training (Tian et al., 2023).

Despite this, the direct recognition and processing of MWEs still have unique values in LMs and LLMs: 1. **MWEs can have unique holistic semantics**: Incorporating MWEs with unique holistic semantics, like "kick the bucket" or "摸鱼" [underwork (actual meaning); touching fish (literal meaning)], can enrich the model's language comprehension capabilities. Although this may not significantly reduce the number of tokens, it allows the model to capture the specific semantics of texts containing these MWEs more directly and accurately. 2. **Some MWEs can reduce the number of types**: In some cases, by appropriately selecting MWEs, it might even be possible to reduce the total number of types. For instance, treating common fixed phrases as single units (like "鹦鹉" [parrot], "乒乓" [ping pong]) might reduce the need for their individual parts.

---

[2] https://www.artfish.ai/p/all-languages-are-not-created-tokenized

[3] https://www.ai21.com/blog/announcing-ai21-studio-and-jurassic-1

Due to the vast amount and diversity of MWEs, there was a scarcity of MWE lexicons. This scarcity consequently hindering their integration into current development of large language models. However, linguists and psycholinguists have long studied on MWEs, and we can rediscover their value based on human cognition:

- **Combining model performance with human language cognition**: With technological advancements, especially when LMs reach engineering limits, LMs can draw more from human language cognition processes.

- **Beyond pure computational power**: Although big data and powerful computing can solve MWE processing to a certain extent, this "brute force" method might not be as efficient and precise as a carefully designed LLM/tokenizer that can directly handle MWEs. Like the convolutional neural network (Lecun et al., 1998) for deep learning and Reinforcement Learning from Human Feedback (RHLF)(Ouyang et al., 2022) for LLMs, we can reconsider drawing inspiration from human language cognition processes when reaching engineering limits.

While the tokenization of subwords and MWEs can offer certain advantages for LMs, exploring deeper principles of language processing is still needed in understanding and optimizing tokenizers. This article will focus an insight of human language acquisition and use - the "Principle of Least Effort".

## Optimizing Future Tokenizers

In the field of engineering, particularly due to the rapid development of NLP and language models, there has been a growing interest in research on tokenizers. This article argue that tokenizers, more than mere technical tools, should emulate and learn from human language processing methods. This argument is based on two main reasons:

- **Language as a cognitively direct product**: While it may be argued that the invention of the aeroplane relied on aerodynamics rather than emulating the way birds fly and suggesting that research on language models and tokenizers need not emulate human cognition, there is a fundamental difference. The aeroplane is designed to fly, not to mimic birds. However, tokenizers directly process content that originates from human cognition, such as written or spoken language. As evidence, BPE-family tokenizers, which are the current dominant tokenization algorithms, also demonstrate their superiority in terms of cognitive plausibility

(Beinborn & Pinter, 2023). In the quest to optimize tokenizer design, therefore, one cannot overlook the mechanisms of human language processing.

- **Lack of a general theory**: After the decline of linguistic-based approaches in the LLM era, the development of tokenizers has often been aimed directly at specific, engineering objectives - such as splitting infrequent words into subwords (e.g., Sennrich et al., 2016) and improving the performance of LMs. This shift has occurred without the foundation of a new, general theoretical framework to guide the shift. This gap has led to reliance on trial-and-error, making the development time-consuming and difficult to provide systematic guidance for subsequent research. In contrast, cognitive science, which has conducted extensive research on tokenization (e.g., Arnon & Priva, 2013; Goldwater et al., 2009; Perruchet & Vinter, 1998; J. Yang, Cai, et al., 2020), is well placed to provide the general theory(s) for the development of NLP tokenizers.

This article will present the "Principle of Least Effort," a general theory from cognitive science that can be applied to tokenizers. It also introduces the Less-is-Better (LiB) model, which is based on this principle, to demonstrate how cognitive science can guide the development of tokenizers.

## Principle of Least Effort

The acquisition of a linguistic token of a human means that the human is able to perceive and generate the token holistically (but can also decompose it if necessary), thereby reducing the complexity of language (Isbilen & Christiansen, 2020). Such **cognitively holistic tokens**, whether they are words, subwords or multiword expressions (Figure 1), can be referred to as the "*cognitive units*" in our mental lexicon[4] (J. Yang, 2022). Furthermore, human tokenization can be described as the process of learning and recognizing these cognitive units. In contrast to the strict definitions of subwords, words, or MWEs, cognitive units are characterized by their adaptability in size and form. This adaptability is evident in how infants and illiterate individuals acquire language - they acquire and use various forms of cognitive units from their environment, even without a formal understanding of what a "word" is. This observation hints that the humans are capable of identify and adopt suitable cognitive units from language inputs autonomously. Current unsupervised tokenization algorithms, which aim to capture linguistic regularities such as frequencies (e.g., Sennrich et al., 2016) and transition probabilities (e.g., Brugnara et al., 1993), reflect this capacity. Yet, despite knowing that the brain can learn these probabilistic regularities (Isbilen et al., 2020; Meltzoff et al., 2009; Schapiro et al., 2016), it is difficult to know the specific "algorithm" for human tokenization.

---

[4] Despite the fact that an individual's cognitive ability and linguistic experience can lead to individual differences in our mental lexicon, the shared language community can still maintain a certain degree of consistency in cognitive units. For example, the word "apple" is a cognitive unit for most English speakers, and "苹果" (Chinese translation of "apple") is a cognitive unit for most Chinese speakers.

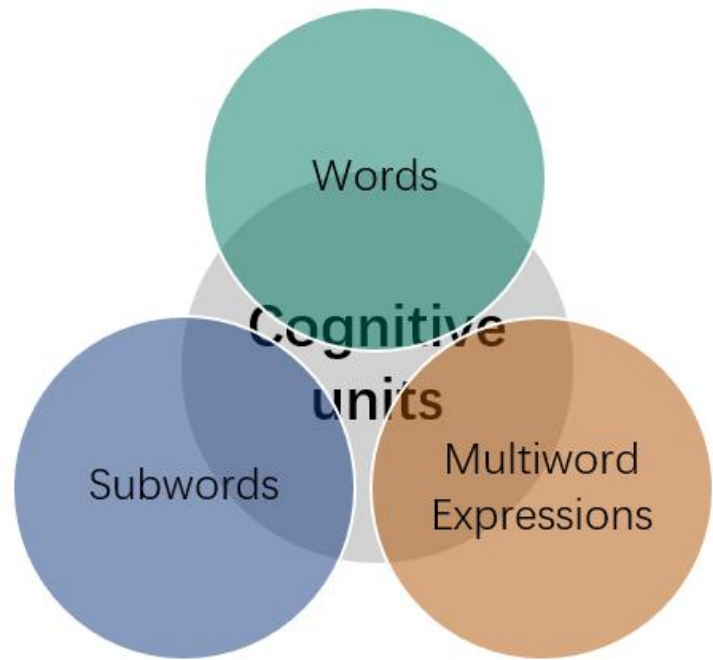

*Figure 1: Some examples of words, adverbs or multi-word expressions that could/are unlikely to become cognitive units in our mental lexicon.*

As a general theory in human cognition, George Kingsley Zipf's Principle of Least Effort (PLE), articulated in his book "Human Behavior and the Principle of Least Effort" (1949), may bridge the gap between the end goal of tokenization and the myriad approaches. This theory is fundamentally a statistical observation about language and other human behavior systems, stating that people tend to follow the path that minimizes effort. One may observe that the description of PLE is simple and the objective of PLE is to be simple, which is also in line with the simplicity principle pursued by many early scholars (e.g., Aristotle, Aquinas, Kant William of Ockham, Newton, and Kant; see: Baker, 2022) and current cognitive scientists (e.g., Chater, 1999; Chater & Vitányi, 2003; Feldman, 2016). Applying PLE in language processing suggests minimizing cognitive burden in language learning and usage. This involves achieving a balance where each cognitive module involved in language processing pursues its own minimal burden. However, language processing is a complex task involving multiple cognitive modules. For the sake of practicality, the process should be simplified.

Zipf's original expression of this principle is encapsulated in his book (1949) that "[the person] will strive to solve his problems in such a way as to minimize the total work that he must expend in solving both his immediate problems and his probable future problems." The mention of "immediate problems" and "probable future problems" are crucial, as they encompass both short-term and long-term needs to reduce burden, which may be conflicting. Thus, following PLE means achieving a balance between short-term and long-term cognitive burden (it is also consistent with compression theory; see the next paragraph). In the previous sections, we have evaluated various approaches based on the number of Types and Tokens, and the benefits of fewer types and fewer tokens have also been demonstrated by LLM experiments (Delétang et al., 2023; Ruoss et al., 2023). From the perspective of PLE: 1. **Fewer tokens can lessen the burden of working memory storage and information decoding steps**; 2. **Fewer types can alleviate the burden of long-term memory storage and retrieval**. This principle can also help us understand the shift from word-level tokenizers to subword-level tokenizers, and supports the introduction of MWEs in the design of more effective tokenizer for LMs.

Moreover, recent studies (Delétang et al., 2023; Gruver et al., 2023) proposed that LMs can be viewed as approximating optimal data compression. This aligns with the Principle of Least Effort. Using Minimum Description Length (MDL) theory (Rissanen, 1978), we can see that fewer types represent a more compressed description of the encoder model, and fewer tokens represent a more compressed description of the encoded data. Compression seeks the minimal total of the encoder model and encoded data. It is worth noting that a crucial difference from this data compression theory is that each brain module seeks its own minimal burden until the global balance is achieved, since the brain's various areas work in coordination and competition, not necessarily managed by a single global controller.

## LiB Model: An Implementation of 'Principle of Least Effort'

In response to the limitations of existing tokenizer technologies in LMs, this section will introduce a new tokenizer design based on PLE. The Least Effort itself, being a principle, can be implemented in various ways. In previous studies (J. Yang, Frank, et al., 2020; J. Yang et al., 2022), the author proposed an implementation focused on reducing the burden of working memory (number of tokens) and long-term memory (number of types), namely the Less-is-Better (LiB) model. This model aims to mimics the learning of the flexible language units. It breaks through the barriers in defining various linguistic units through unsupervised methods, and unifies subwords, words, and multi-word expressions (MWEs) into the same vocabulary. In this process, it effectively balances the number of tokens and types to reduce the cognitive burden of using language (Figure 2). The process is, to some extent, approaching to the Minimum Description Length, but with a focus on the balance between two individual minimization (min(#tokens) vs. min(#types)) rather than a global minimization (min(#tokens + #types)).

**Model Mechanism**: The model consists of a "Memorizer" and a "Forgetter". Initially, the LiB model splits the input corpus into the smallest tokens and then the "Memorizer" continuously merges adjacent tokens in the corpus into new (longer) units and stores them in the vocabulary. By using longer units, the number of unit tokens in the text decreases, while the number of types increases. Conversely, the "Forgetter" removes less useful "junk" units from the vocabulary to reduce the number of unit types. "junk" units may be those types that increase the number of unit tokens in sentences or are infrequently appearing types. The "Memorizer" and "Forgetter" balance each other, eventually reaching a relatively steady state, where the vocabulary contains units close to the goal of cognitive burden minimization. See more detail in J. Yang, Frank, et al., (2020).

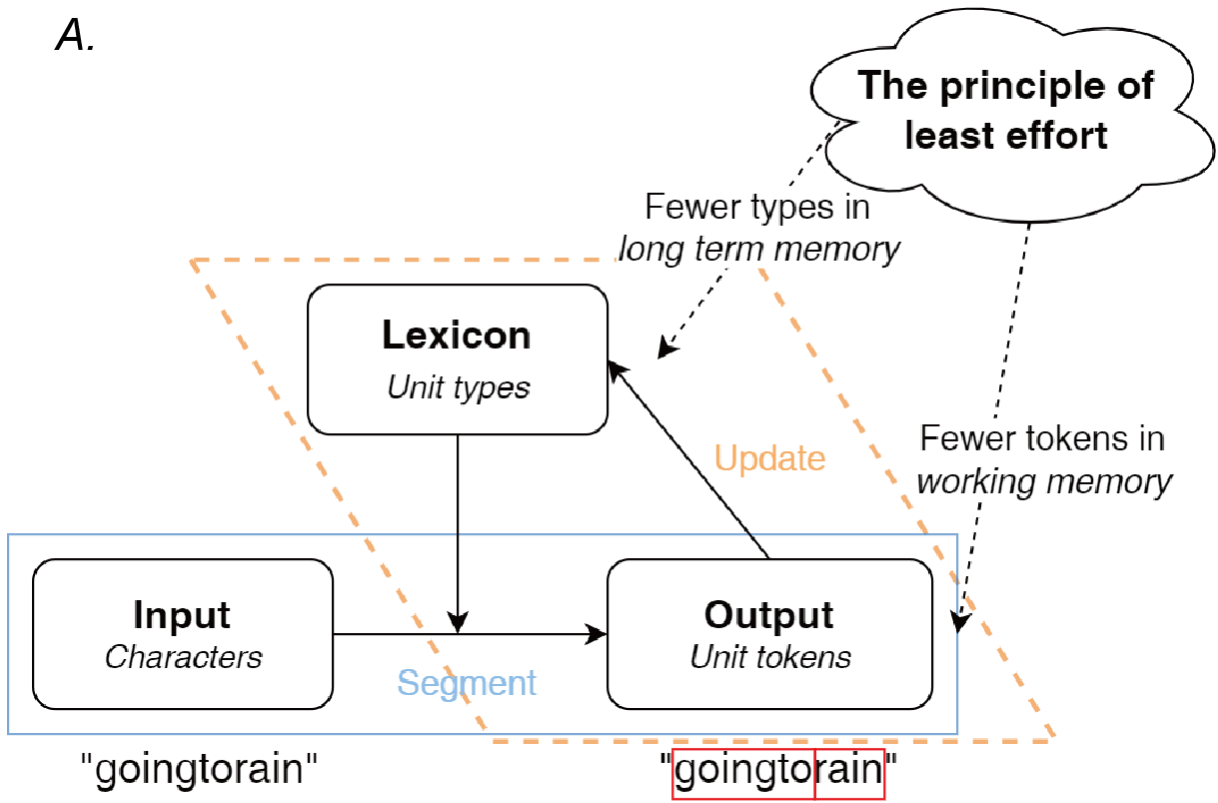

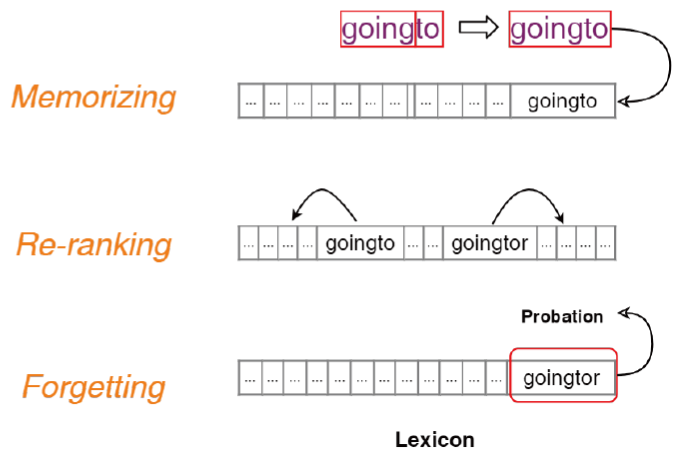

Figure 2: The LiB model. A. the overview information flow, B. the strategies of text segmentation, C. the strategies of lexicon/vocabulary update.

**Results:** The LiB model's unsupervised method ignores the definition barriers between various traditional linguistic units, so its vocabulary also breaks through the usual limitations of subwords, words, and multi-word expressions. Two sentences are presented in Table 4 for the demonstration. On the vocabulary level, the model autonomously learns variform English units like "ly," "you," and "you can", as well as Chinese units like "的" (English translation: "'s"), "孩子" (English translation: "kid"), and "新华社" (English translation: "Xinhua News Agency") (J. Yang, Frank, et al., 2020). This fusion reflects the LiB model's flexibility in learning cognitive units of different sizes and linguistic levels.

| Corpus | Type | Segmentation |
| --- | --- | --- |
| BRphono | Input | allrightwhydon'tweputhimawaynow |
|  | Words | all\|right\|why\|don't\|we\|put\|him\|away\|now |
|  | LiB output | allright\|whydon't\|we\|puthimaway\|now |
| CTB8 | Input | 这个出口信贷项目委托中国银行为代理银行 |
|  | Words | 这\|个\|出口\|信贷\|项目\|委托\|中国\|银行\|为\|代理\|银行 |
|  | LiB output | 这个\|出口信贷\|项目\|委托\|中国银行\|为\|代理\|银行 |

*Table 4: Example segmentations of strings in the two corpora. BRphono's results are transcribed into English words for ease of presentation (see Table 3 in J. Yang, Frank, et al., (2020)).*

**Practical Application**: The units learned by LiB can be used to predict the eye fixation patterns of human readers, suggesting that the model's units are consistent with human cognitive units (J. Yang et al., 2022). For corpora in different languages, the LiB model flexibly learns their lexicons through the unsupervised method based on PLE (J. Yang, Frank, et al., 2020; J. Yang et al., 2022), thereby adapting to the complexity and diversity of different language inputs, while balancing cognitive loads. Although LiB is only a cognitive model and has not been optimized for language models, evaluations on simple language models show that LiB-generated units perform better in Bits-per-character scores (Table 5). This superior performance may be attributed to that LiB learns fewer tokens and types than word-level tokenizers and BPE tokenizers (Table 6). This suggests the value of PLE in this era of LARGE language models. We may use the LiB model or other variants that also follow PLE as tokenizers for large language models to enhance their performance.

| Corpus | Metric | Tokenizations | | | |
|---|---|---|---|---|---|
| | | Characters | BPE subwords | Words | LiB units |
| BRphono (English) | Average token length | 1 | 2.8 | 2.9 | 3.6 |
| | Vocabulary size | 50 | 5,574 | 1,321 | 1,869 |
| CTB8 (Chinese) | Average token length | 1 | 1.4 | 1.7 | 1.9 |
| | Vocabulary size | 4,697 | 7,980 | 65,410 | 39,320 |

Table 5: Average token lengths, lexicon sizes of different tokenizations on the two corpora. The unit of Average Length is English phoneme (BRphono) or Chinese character (CTB8) (see Table 4 in J. Yang, Frank, et al., (2020)).

| Corpus | Model | Character | BPE subword | Word | LiB chunk |
|---|---|---|---|---|---|
| CTB8 | 2-Gram | 3.558 | 2.788 | 2.333 | **2.095** |
| | 3-Gram | 2.025 | 1.193 | 1.163 | **0.903** |
| BRphono | 2-Gram | 2.221 | 1.003 | 0.977 | **0.791** |
| | 3-Gram | 1.371 | 0.563 | 0.584 | **0.484** |

Table 6: Bits per character scores on different tokenizations (see Table 5 in J. Yang, Frank, et al., (2020)).

This cognitive science-based approach provides a new perspective and direction for the future development of language models, especially when dealing with corpora in various languages (like Chinese, which lacks clear word boundaries).

## Summary

This article explores the current choice and future optimization of tokenizers for large language models (LLMs), especially in handling complex languages like Chinese. Overall, subword tokenization, as a balancing technique, significantly reduces the number of types while only slightly increasing the number of tokens compared to word tokenization, effectively addressing Out-Of-Vocabulary (OOV) issues and enhancing the model's generalization capabilities. However, this method has limitations in controling the number of tokens in some non-Latin languages (like Chinese), and also in capturing the nuanced semantics and idiom implications of language.

The absence of MWEs in most LMs reflects a blind spot in the current NLP field. Although MWEs significantly increase the number of types, and current models can learn the meanings of MWEs on subwords tokenization by massive data/computational power, direct recognition and processing of MWEs can still help language models improve the accuracy in language understanding. In future development of tokenizers, how to effectively select MWEs and balance the number of tokens and types could be a key area for tokenizer advancement.

To address the issues of current tokenizer technologies, this article discussed the importance of emulating human language processing methods in tokenizer design, and introduced the "Principle of Least Effort" from cognitive science, which not only reveals efficiency and simplicity in human language processing but also, as a general theory in cognitive science, can guide the design of more efficient tokenizers. Based on this principle, this article proposed the LiB model, a model that attempts to optimize its vocabulary through learning and forgetting mechanisms, achieving a more effective balance of tokens and types. It aims to simulate human language processing mechanisms, reducing cognitive burden, and obtaining new types of linguistic cognitive units that integrate subwords, words, and MWEs, thereby enhancing the efficiency and accuracy of language processing. The LiB model is not only a reflection on human language processing mechanisms but also provides new ideas for designing more effective tokenizers for LMs. This cognitive science-based approach provides new perspectives and directions for the future development of tokenizers and language models. Incorporating insights from cognitive science with the design of large language models may enhance their synergistic evolution.

## Acknowledgements

JY was supported by the Lise Meitner Research Group "Language and Computation in Neural Systems" of Dr. Andrea E. Martin, funded by the Max-Planck Society and the Max-Planck-Institute for Psycholinguistics.